\documentclass[sigconf]{acmart}

\renewcommand\footnotetextcopyrightpermission[1]{} 
\usepackage{booktabs} 
\usepackage{multirow}
\usepackage{amsmath}
\usepackage{color}
\usepackage{arydshln }

\acmConference[WSDM]{The Twelfth International Conference on Web Search and Data Mining}{February 11--15}{Melbourne, Australia} 
\acmYear{2019}

\begin{document}
\title[Missing Information Loss]{A Missing Information Loss for implicit feedback datasets}

\author{Juan Ar\'evalo}
\affiliation{%
  \institution{BBVA Data \& Analytics}
}
\email{juanmaria.arevalo@bbvadata.com}

\author{Juan Ram\'on Duque}
\affiliation{%
  \institution{BBVA Data \& Analytics}
}
\email{juanramon.duque@bbvadata.com}

\author{Marco Creatura}
\affiliation{%
  \institution{BBVA Data \& Analytics}
}
\email{marco.creatura@bbvadata.com}

\renewcommand{\shortauthors}{Ar\'evalo, Duque and Creatura}

\newcommand{\MFsquare}{\textsc{MF-square}}
\newcommand{\MFmil}{\textsc{MF-mil}}
\newcommand{\MFce}{\textsc{MF-CE}}
\newcommand{\CEpointlinsig}{\textsc{CE$_{\rm Point}$ lin-sig}}
\newcommand{\CEpointsigsig}{\textsc{CE$_{\rm Point}$ sig-sig}}
\newcommand{\CEpairlinsig}{\textsc{CE$_{\rm Pair}$ lin-sig}}
\newcommand{\CEpairsigsig}{\textsc{CE$_{\rm Pair}$ sig-sig}}
\newcommand{\MULTItanhlin}{\textsc{MULTI tanh-lin}}
\newcommand{\MILlinsig}{\textsc{MIL lin-sig}}
\newcommand{\MILsigsig}{\textsc{MIL sig-sig}}

\begin{abstract}
Latent factor models for Recommender Systems with implicit feedback typically treat unobserved user-item interactions (\emph{i.e.} missing information) as negative feedback. This is frequently done 
either through negative sampling (point--wise loss) or with a ranking loss function (pair-- or list--wise estimation). 
Since a zero preference recommendation is a valid solution for most common objective functions, 
regarding unknown values as actual zeros results in users 
having a zero preference recommendation  for most of the available items. 

In this paper we propose a novel objective function, the \emph{Missing Information Loss} (MIL), 
that explicitly forbids treating unobserved user-item interactions as positive or negative feedback. 
We apply this loss to both traditional Matrix Factorization and user--based Denoising Autoencoder, and compare it with other established objective functions such as cross--entropy (both point-- and pair--wise) or the recently proposed multinomial log-likelihood. MIL achieves competitive performance in ranking--aware metrics when applied to three datasets.
Furthermore, we show that such a relevance in the recommendation is obtained while displaying popular items less frequently (up to a $20 \%$ decrease with respect to the best competing method). This debiasing from the recommendation of popular items favours the appearance of infrequent items (up to a $50 \%$ increase of long--tail recommendations), a valuable feature for Recommender Systems with a large catalogue of products. 
\end{abstract}

\ccsdesc[500]{Information systems~Recommender systems}

\keywords{Collaborative Filtering, Autoencoders, Implicit Feedback, Missing Information}

\setcopyright{None}

\maketitle

\setlength{\abovecaptionskip}{0pt}
\setlength{\belowcaptionskip}{-10pt}


\section{Introduction}\label{sec:intro}

Nowadays, users are faced with such a large volume of  products and information to the extent that filtering has become a necessity. 
Furthermore, not every user has the same preferences, and therefore a standard selection cannot be performed.
Both problems can be tackled with Recommender Systems (RS), that provide users with a personalized list of items.
Moreover, for organizations with vast inventories, it is of great interest to recommend infrequent products, \emph{i.e.} items queuing in the long tail of the catalogue of items~\cite{Anderson:2006:long_tail}.  
Nevertheless, most RS algorithms tend to over-recommend popular items, precisely because their metric performance generally decreases when recommending less frequent items~\cite{Steck:2011:IPR}.   

An increasingly adopted approach to RS is collaborative filtering (CF) for implicit feedback datasets~\cite{HuKoren:2008:CF_implicit, Pan:2008:OCCF}.
This technique makes predictions about the interests of a user
by gathering  preferences in the form of purchases, clicks, logs, etc. from many other users.
These preferences are influenced by non-obvious factors that depend on the domain.
The goal of model based CF approaches such as matrix factorization is to infer
the latent factor model underlying the data.

Making predictions about user preferences in RS based on implicit feedback data is not an easy task, not only because of  the lack of information about unobserved user-item interactions, but also due to the subsequent high sparsity of the rating matrix.
In order to deal with this issue of missing information, several approaches can be considered. 
A naive strategy for one--class collaborative filtering consists of considering all unobserved items either as negative examples (All Missing As Negative) or simply to ignore them (All Missing As Unknown)~\cite{Pan:2008:OCCF}. However, these two extreme methodologies may involve either biased recommendations (as some of the missing data might be relevant to the user) or trivial solutions caused by predicting all missing values as positive examples.
More advanced approaches entail the use of different weighting schemes in the error terms to balance observed and unobserved items~\cite{HuKoren:2008:CF_implicit, Pan:2008:OCCF, Steck:2010:MissingNotAtRandom}. In order to alleviate the computational burden of considering every single item during the training process  (which does not scale linearly as the item catalogue grows), several negative sampling techniques have been proposed, see \emph{e.g.}~\cite{Pan:2008:OCCF}. 

It should be noted that all these approaches cast missing entries as zeros, \emph{i.e.} treat unobserved user-item interactions as negative feedback. Since the loss functions considered in those works are designed to push items with negative feedback towards a zero recommendation, in practice  a zero preference will be inferred for most of the missing entries. 

In addition to the one-class collaborative filtering setting, the RS problem can be viewed as a multi-class classification problem, where the recommendation problem is cast as the calculation of the probability of a user belonging to each item class. Indeed, the multinomial distribution has been recently applied to different Autoencoder (AE) architectures~\cite{liang:2018:VAE}. In contrast to the above-mentioned objective functions, the log-likelihood of the multinomial distribution does not explicitly penalize missing entries so as to force them to have a zero probability of being recommended. However, the normalization condition of the probability distribution, together with the large item catalogues used in RS (typically  $>10$k items), make it unlikely that non-seen items have a probability of being recommended other than zero.

In this paper we propose to tackle the problem of missing information by using a novel objective function that specifically forbids missing user--item interactions to have a preference prediction of $1$ (since they are not positive feedback) or $0$ (to distinguish them from negative feedback).  We name this cost function the \emph{Missing Information Loss} (MIL).
We apply MIL to both Matrix Factorization (ML) and Denoising AE (DAE)~\cite{Vincent:2008:ECRF-AE}, as the later has been shown to be a generalization of traditional MF models~\cite{Wu:2016:CDAE-topN}. We compare the results with DAEs optimized with cross-entropy loss (in both point and pair--wise learning schemes)~\cite{Wu:2016:CDAE-topN} and the multinomial loss~\cite{liang:2018:VAE}; as well as MF models~\cite{HuKoren:2008:CF_implicit} trained with square and cross-entropy loss. 

Our contributions can be summarized as follows:
\begin{itemize}
\item We propose a new objective function (MIL) for modeling missing information in implicit feedback datasets. It explicitly forbids either a $1$ or $0$ prediction for the preference of unobserved user--item interactions, thus leaving the ranking process almost entirely to the low-rank process underlying all forms of matrix factorization. 
\item We show that the \textsc{MIL} function achieves state-of-the-art metric performance when applied to MF and DAE architectures, 
similar to the best performing, well established, objective functions. 
\item Furthermore, we demonstrate that the observed competitive performance (in terms of relevance) occur while recommending popular items less frequently, which favours the appearance of medium and long--tail items in the ranked list of recommendations.
\end{itemize}
	
The rest of the paper is organized as follows.
In section \ref{sec:model} we review some of the objective functions that are commonly used in RS literature, and introduce the MIL function. Next, we briefly revisit the DAE and MF architectures. 
After that, in section \ref{sec:protocols}, we describe the experimental methodology: datasets, metrics for evaluation, the implementation details of the proposed solution, and the baseline models used for comparison.
Next, section \ref{sec:results} shows the experimental results in terms of ranking metrics, as well as the observed distribution of recommendations.
Finally, we draw some conclusions in section \ref{sec:conclusions} and 
indicate future lines of research.

\section{Model}\label{sec:model}
Let $\mathcal{U}$ be the set of users and $|\mathcal{U}|$ the total number of users. 
Let $\mathcal{I}$ be the set of available items and $\mathcal{I}_u$ the set adopted by user $u$. 
The number of total items is denoted by $|\mathcal{I}|$, and those adopted by a user $|\mathcal{I}_u|$.
Our goal is to predict the preference of a user for all the available items given its history of binary preferences $p_{ui} \mid i\in\mathcal{I}_u$, i.e. calculate $\hat{p}_{ui}~\forall i\in \mathcal{I}$.
Next, we review some familiar objective functions, and introduce our Missing Information Loss (MIL) function. Then, we revisit the user--based Denoising Autoencoder (DAE)~\cite{Kramer:1991:NLPCA, Vincent:2008:ECRF-AE} and specify its usage for RS~\cite{Sedhain:2015:Autorec, Wu:2016:CDAE-topN}.

\subsection{Objective functions for Recommender Systems}\label{subsec:losses}

Learning to assign user preferences for items depends to a great extent on how the objective function--i.e., the function we intend to optimize--is set. In this subsection we review the most relevant objective functions considered in the literature for the task of recommendation, with a special focus on the missing information issue.

Regarding the square loss, a confidence scale factor for balancing the observed and unobserved items is introduced in~\cite{HuKoren:2008:CF_implicit}.
This factor can be defined as $C(p_{ui}):=a p_{ui}$, with $a>1$ a hyper-parameter for ensuring a correct balance. 
Using such confidence scale factor, the square loss is cast as
\begin{equation}\label{eq:square}
l(p_{ui},\hat{p}_{ui})=\frac{C(p_{ui})+1}{2}\left(p_{ui}-\hat{p}_{ui}\right)^2.
\end{equation}
Similarly, the cross-entropy objective function can be generalized to account for the unbalance of classes, 
\begin{equation}\label{eq:cross-entropy}
l\left(p_{ui},\hat{p}_{ui}\right) = 
- C\left(p_{ui}\right)\log\left(\hat{p}_{ui}\right)  
- \left(1-p_{ui}\right)\log\left(1-\hat{p}_{ui}\right).
\end{equation}
In both cases, the total loss is averaged across all users, 
\begin{equation}\label{eq:point-wise_no_sampling}
\mathcal{L}_{\rm point}=\frac{1}{|\mathcal{U}|}\sum_{u\in \mathcal{U}} 
\sum_{i\in\mathcal{I}} l\left(p_{ui},\hat{p}_{ui}\right).
\end{equation}
Please note that casting unobserved user-item interactions as  $p_{ui}=0$ in equations (\ref{eq:square}) and (\ref{eq:cross-entropy}), induces many zero recommendations, \emph{i.e.} $\hat{p}_{ui}=0$. With all certainty, the limited capacity of the model (the low--rank process) avoids setting all unobserved items with a zero preference prediction. 

Due to the large item catalogues typically involved in RS, negative sampling techniques are used to solve  the positive/negative class unbalance problem~\cite{Pan:2008:OCCF, Wu:2016:CDAE-topN}. For this, a target set $\mathcal{T}_u$ is built by joining the observed item set $\mathcal{I}_u$ and items sampled from $\mathcal{\tilde{I}}_u:=\mathcal{I} \setminus \mathcal{I}_u$. The number of items sampled from $\mathcal{\tilde{I}}_u$ is a hyper-parameter to be tuned, while $C(p_{ui})$ in equations (\ref{eq:square}) and (\ref{eq:cross-entropy}) is set to $1$  for all preferences. The loss is then computed as 
\begin{equation}\label{eq:point-wise}
\mathcal{L}_{\rm point}=\frac{1}{|\mathcal{U}|}\sum_{u\in \mathcal{U}} 
\sum_{i\in\mathcal{T}_u} l\left(p_{ui},\hat{p}_{ui}\right).
\end{equation}

The above objective functions are examples of point-wise learning, where the loss is calculated by taking the information of only one item at a time. Rendle \emph{et al.}~\cite{Rendle:2009:BPR} introduced 
pair-wise learning, which confronts a pair of items (positive and  unobserved) to compute the final loss. 
Because of this, a new set $\mathcal{P}_u$  consisting of pairs of seen (positive feedback) and unseen items  (assumed negative feedback) is created. The total pair-wise loss is then defined as~\cite{Wu:2016:CDAE-topN}
\begin{equation}\label{eq:pair-wise}
\mathcal{L}_{\rm pair}=\frac{1}{|\mathcal{U}|}\sum_{u\in \mathcal{U}} 
\sum_{i,j\in\mathcal{P}_u} l\left(p_{uij},\hat{p}_{uij}\right).
\end{equation}
Here, $p_{uij}:=p_{ui}-p_{uj}=1,\forall (i,j)\in\mathcal{P}_u$ and $\hat{p}_{uij}:=\hat{p}_{ui}-\hat{p}_{uj}$. 

For both point and pair--wise learning schemes, the objective functions defined in equations~(\ref{eq:square}) and (\ref{eq:cross-entropy}) admit as a valid solution a predicted zero preference 
when the input preference is zero, i.e. $\hat{p}_{ui}=0$ if $p_{ui}=0$. However, it should be noted that in implicit feedback datasets there are no actual zero preferences, but rather missing information. Thus, by using any of the losses described above the  solution will inevitably assign zero preferences to most of the unobserved user-item pairs.  This fact affects the way in which items are recommended, as discussed in section~\ref{sec:results}.

On the other hand, a model based on the multinomial distribution has been recently applied to AEs by Liang \emph{et al.}~\cite{liang:2018:VAE}. The log-likelihood for a user $u$ in this setting can be written as
\begin{equation}\label{eq:multinomial}
-\sum_i p_{ui}\log\pi_i\left(\hat{p}_{ui}\right)
\end{equation}
where $\pi_i\left(\hat{p}_{ui}\right)$ is the probability distribution of the predictions.
Note that in contrast to the square and cross--entropy losses, this objective function does not explicitly penalize missing values, since $p_{ui}=0$ for unobserved user--item interactions.
Instead, the normalization condition of the probability distribution  ($\sum_i\pi_i\left(\hat{p}_{ui}\right)=1$), together with the low--rank process, helps to assign non-zero preferences to the unobserved items. However, the large item catalogues used in RS (typically  $>10$k) make it unlikely that non-seen items have a probability different from zero. Furthermore, the normalization condition on the  probabilities prevent this modeling from scaling up.

\begin{figure}[hbt]
    \centering
    \includegraphics[width=.75\linewidth]{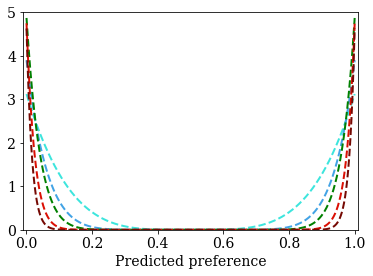}
    \caption{Different values of the hyper-parameters $\boldsymbol{A_{\rm MI}}$ and $\boldsymbol{\gamma_{\rm MI}}$ modeling the missing information term in equation~(\ref{eq:mil_def}). The exponent of each polynomial here is, from the smoothest to the sharpest curves, $\boldsymbol{2 \gamma_{\rm MI}\equiv\{4,8,12,20,30\}}$.
    }
    \label{fig:mil_parameters}
\end{figure}

In order to mitigate all the problems mentioned within this subsection, we propose a novel objective function, the \emph{Missing Information Loss} (MIL), that explicitly forbids treating missing information as positive or negative feedback. For this reason, we propose the functional form
\begin{eqnarray}\label{eq:mil_def}
l(p_{ui},\hat{p}_{ui}) &=& 
\frac{1}{2} p_{ui} (1 + p_{ui}) (1 - \hat{p}_{ui})^{\gamma_{+}} + \nonumber\\
             & & \frac{1}{2} (1+p_{ui})(1-p_{ui}) A_{\rm MI}(\hat{p}_{ui}-0.5)^{2\gamma_{\rm MI}}. 
\end{eqnarray}
Here, the first and second term evaluate the contribution of the observed and unobserved user-item pairs into the final loss.
In particular, the first term estimates preferences for positive items as a power law with parameter $\gamma_+$. Indeed, for $p_{ui}=1$ equation~(\ref{eq:mil_def}) reduces to
\begin{equation*}
l(p_{ui}=1,\hat{p}_{ui}) = (1 - \hat{p}_{ui})^{\gamma_{+}}. 
\end{equation*}
On the other hand, the last term in (\ref{eq:mil_def}) explicitly forbids predicted 0 and 1 preferences for missing entries, acting as a barrier for the optimization process. As a matter of fact, for $p_{ui}=0$ equation~(\ref{eq:mil_def}) is cast as
\begin{equation*}
l(p_{ui}=0,\hat{p}_{ui}) = A_{\rm MI}(\hat{p}_{ui}-0.5)^{2\gamma_{\rm MI}}. 
\end{equation*}
The constants $A_{\rm MI}$ and $\gamma_{\rm MI}$ are hyper-parameters to be fine-tuned. In this paper we explore the pairs $\left(A_{\rm MI}, \gamma_{\rm MI}\right)\in\{
(5\cdot10^1, 2)$, $(10^3, 4)$, $(2\cdot10^4, 6)$, $(1\cdot10^6, 10)$, $(5\cdot10^9, 15)
\}$, see Figure~\ref{fig:mil_parameters}.

At this point, it is worth stressing that under the MIL function all items--independently of their position into the long tail curve--can be part of the recommendation process, as their predicted preferences adopt non-zero values, $\hat{p}_{ui}\in(0,1)$.
Thus, the final predicted preference will be adjusted by the collaborative filtering among users, 
the co-occurrence of items, and the limited capacity of the model--\emph{i.e.}, the overall low--rank process.

Note that the MIL function in equation~(\ref{eq:mil_def}) can be naturally extended to account for actual negative feedback, \emph{i.e} $p_{ui}=-1$. Indeed, we can simply add the term 
\begin{equation}\label{eq:neg_feedback_term}
    -\frac{1}{2} p_{ui} (p_{ui}-1) \hat{p}_{ui}^{\gamma_{-}}. 
\end{equation}
to equation~(\ref{eq:mil_def}), which vanishes whenever $p_{ui}=0,1$.
Here, $\gamma_{-}$ is an exponent controlling the family of polynomials modeling negative feedback entries. Such a term would force negative ratings to have a zero predicted preference. 
We will leave the analysis of datasets with actual negative feedback for future study.

Finally, for any given loss function we regularize the model with weight-decay, so that the total loss is 
\begin{equation}\label{eq:L2_reg}
\mathcal{L} = \mathcal{L}_{\rm point/pair}+\lambda\left( ||{\bf W}||_2^2+||{\bf W'}||_2^2\right).
\end{equation}
Here, $\lambda$ is a hyper-parameter.

\subsection{Architecture details}\label{subsec:architecture}
We represent an  item $i$ is as a one-hot encoding ${\bf v}_i$, i.e. a $|\mathcal{I}|$-sized vector of zeros with a 
$1$ at position $i$. 
Next, we represent a user $u$ as the sum of the one-hot encoding vectors of items in $\mathcal{I}_u$,
\begin{equation}\label{eq:user_input}
{\bf v}_u: =\sum_{i\in\mathcal{I}_u}{\bf v}_i = \sum_{i\in\mathcal{I}}p_{ui}{\bf v}_i,
\end{equation}
where $p_{ui}$ is the preference of user $u$ for item $i$, i.e. $p_{ui}=1$ if $i\in\mathcal{I}_u$, $p_{ui}=0$ otherwise.
The vector of preferences is therefore equal to the user vector in our setting, ${\bf p}_u\equiv {\bf v}_u$.
In the general case of non binary implicit ratings $r_{ui}$ (such as purchases or play counts) 
the user vector will consist of the weighted sum of the one-hot encoding of their items,
\begin{equation}\label{eq:user_input_weighted}
{\bf v}_u: =\sum_{i\in\mathcal{I}_u}r_{ui}{\bf v}_i.
\end{equation}

An Autoencoder (AE) \cite{Kramer:1991:NLPCA} is a feedforward neural network for learning a representation 
of the input data. This representation is trained to produce an output that closely matches the original input. 
When applied to RS, the encoder typically has  a much lower dimension than the input vector; 
hence, the learned representation has to encode input information while reducing the dimensionality of the input space. 
In this paper we consider a single hidden layer AE.
The input vector of preferences, ${\bf p}_u\in\mathbb{R}^{|\mathcal{I}|}$, is projected onto a vector ${\bf h}\in\mathbb{R}^D$,
\begin{equation}\label{eq:AE_hidden}
{\bf h} = s({\bf W}{\bf v}_u+{\bf b}).
\end{equation}
Here ${\bf W}\in\mathbb{R}^{D\times|\mathcal{I}|}$  and ${\bf b}\in\mathbb{R}^D$ are learnable weight matrix and bias vectors, respectively. 
The activation $s(\cdot)$ is an element-wise mapping function; typical activations are the sigmoid function, the hyperbolic tangent or
the Rectified Linear Unit (ReLU)~\cite{icml2010_NairH10_Relu}. 

In order to obtain the predicted preferences, the hidden layer is projected back onto the original space, 
\begin{equation}\label{eq:AE_output}
\hat{\bf p}_u = s'({\bf W}'{\bf h}+{\bf b}'),
\end{equation}
where ${\bf W}'\in\mathbb{R}^{|\mathcal{I}|\times D}$  and ${\bf b}'\in\mathbb{R}^{|\mathcal{I}|}$ are weight matrix and bias vectors for the output layer,
and $s'(\cdot)$ is the activation of the decoder (which may or may not be equal to that used when encoding). 
The vector of predicted preferences, $\hat{\bf p}_u$, is then forced to minimize the objective functions defined in subsection~\ref{subsec:losses}. 
A variant of the AE is the Denoising Autoencoder (DAE) \cite{Vincent:2008:ECRF-AE}, which attempts to
reconstruct a corrupted version of the input $\tilde{{\bf x}}$, i.e. ${\bf h} = s({\bf W}\tilde{{\bf x}}+{\bf b})$. This latter technique has become very popular due to its success in image recognition, and is currently applied in most AEs for RS~\cite{Wu:2016:CDAE-topN, liang:2018:VAE}. 

Please note that for MIL and cross-entropy losses (equations~(\ref{eq:mil_def}) and (\ref{eq:cross-entropy}) respectively), output preferences must be bounded $\hat{p}_{ui}\in(0, 1)$. Thus, we typically choose the sigmoid function for the activation of the decoder, $s'=\sigma$. This is different from the also traditional approach of minimizing the logistic log-likelihood~\cite{Wu:2016:CDAE-topN, liang:2018:VAE}, which already incorporates the sigmoid function into the loss, and thus allows one to apply yet another activation at the decoder (e.g. $\tanh(\sigma(\hat{p}_{ui}))$, as in \cite{liang:2018:VAE}, or $\sigma(\sigma(\hat{p}_{ui}))$ in \cite{Wu:2016:CDAE-topN}). 
\paragraph{Matrix Factorization}

As shown in~\cite{Wu:2016:CDAE-topN}, the AE described in equations (\ref{eq:user_input}), (\ref{eq:AE_hidden}) and (\ref{eq:AE_output}) is a generalization of Matrix Factorization (MF) models. Indeed, MF is recovered after replacing the input user vector (\ref{eq:user_input}) by the one-hot vector of the user id $u$, which is a $|\mathcal{U}|$--sized vector of zeros with a 1 at the position $u$. The bi--linear MF has linear activations $s$ and $s'$, and is typically trained with square loss and weight--decay regularization, see equations~(\ref{eq:square}), (\ref{eq:point-wise}) and (\ref{eq:L2_reg}).

However, due the the large disparity in the number of users and items, $|\mathcal{U}|\ll |\mathcal{I}|$, the norms of the weight matrices ${\bf W}\in\mathbb{R}^{D\times|\mathcal{U}|}$ and ${\bf W'}\in\mathbb{R}^{|\mathcal{I}|\times D}$ are quite different. Thus, direct application of equation~(\ref{eq:L2_reg}) will tend to over-regularized the user matrix ${\bf W}$, leaving the item one ${\bf W'}$ under-regularized; this may lead to instabilities and potential over-fitting while training (as we have observed experimentally). This issue can be solved by re-scaling the norms of weight matrices in equation~(\ref{eq:L2_reg}) as
\begin{equation}\label{eq:l2_reg_scaled}
\mathcal{L} = \mathcal{L}_{\rm point/pair}+\frac{\lambda}{D}\left( 
\frac{||{\bf W}||_2^2}{|\mathcal{U}|}+
 \frac{||{\bf W'}||_2^2}{|\mathcal{I}|}
\right).    
\end{equation}
This way, both user and item embeddings are regularized with the same strength, stabilizing the training procedure. Note: in the case of an AE, there is no need to re-scale the norm of weights matrices, since both have the same number of elements, $|\mathcal{I}|\times D$.

\section{Experimental protocols}\label{sec:protocols}
\subsection{Datasets}

We use the 
MovieLens--20M\footnote{http://grouplens.org/datasets/movielens}
and Netflix\footnote{http://www.netflixprize.com} explicit feedback  datasets.
As both  of these contain explicit ratings, we create binary preferences
by keeping ratings $\ge\!4$, which we interpret as positive feedback ($p_{ui}=1$).
Furthermore, we only keep users with at least 5 views.
Validation and test sets are obtained randomly, selecting a $10~\%$ of the original dataset for each set. We denote such datasets \textsc{ML20M} and \textsc{Netflix}.

In addition, we explore models performance on the Last.fm\footnote{https://www.upf.edu/web/mtg/lastfm360k} dataset~\cite{Celma:Springer2010}, an implicit feedback dataset consisting of tuples (user, artist, plays), that contains top artists by user. In order to make the comparison with the above datasets more straightforward, we binarize play counts and interpret them as implicit preference data. Next, we filter out artist with less than 50 distinct listening users, and user with less than 20 artists in their listening history. In the following, we name this dataset \textsc{Lastfm}.

\setlength{\belowcaptionskip}{5pt}
\begin{table}[htb]
\begin{tabular}{c c c c c}
 Dataset & \#users & \#items & \#pairs & \#pairs$_{\rm test}$ \\
\hline
\textsc{ML20M} & 136,7k & 20,3k & 7,99M & 1,0M \\
\textsc{Netflix}  & 463,4k & 17,7k & 45,5M & 5,7M \\
\textsc{Lastfm}  & 350,2k & 24,6k & 12,8M & 1,6M \\
\hline
\end{tabular}
\caption{Statistics of the datasets after preprocessing.}
\label{table:datasets}
\end{table}

The statistics of the training set after such  processing, as well as the number of user-item interactions in test, are presented in Table~\ref{table:datasets}. 


\subsection{Evaluation metrics}\label{subsec:metrics}
\setlength{\belowcaptionskip}{-10pt}
Given the set of adopted items in test, $\mathcal{I}_u^{\rm t}$, and the ranked list of predicted preferences, 
the relevance of a recommendation at position $k$ is given by ${\rm rel}_{ui}(k)$--${\rm rel}(k)$ from here on--, which equals $1$ if user $u$ adopted item $i$ in the test set, $0$ otherwise. In the calculation of metrics, we remove items observed in  training and validation from the list of recommendations. 
Next, we detail the metrics used for model evaluation.

\paragraph{Recall} It does not account for the relative ordering of the recommendation, and we defined it as~\cite{liang:2018:VAE}
\begin{equation}
{\rm Recall}@k = \frac{\sum_{s=1}^k {\rm rel}(s)}
{\mathcal{N}_u(k)}.
\end{equation}
Here, $\mathcal{N}_u(k) = \min\left(k,|\mathcal{I}_u^{\rm t}|\right)$, with $|\mathcal{I}_u^{\rm t}|$ the number of items adopted by user $u$ in testing. The final recall is averaged across all users in testing. 

\paragraph{Normalized Discount Cumulative Gain} In contrast to recall metric, the Discount Cumulative Gain (DCG) performs a logarithmic discount according to the position of a recommendation, that is
\begin{equation}
{\rm DCG}@k = \sum_{s=1}^k \frac{{\rm rel}(s)}
{\log_2(s+1)}.
\end{equation}
This quantity can be normalized by the Ideal DCG, 
\begin{equation}
{\rm IDCG}@k = \sum_{s=1}^{\mathcal{N}_u(k)} \frac{1}{\log_2(s+1)}.
\end{equation}
Finally, NDCG$@k= {\rm DCG}@k/{\rm IDCG}@k$, which we average across all users in the test set.

\paragraph{Novelty} Following reference~\cite{Vargas:2011:Novelty_diversity}, we define a novelty-weighted DCG score as
\begin{equation}\label{eq:nov-ndcg}
{\rm Nov\!-\!DCG}@k = 
-\sum_{s=1}^k \frac{{\rm rel}(s)\times \ln{\nu(i)}}
{\log_2(s+1)}.
\end{equation}
Here, $\nu(i)$ is the frequency of occurrences of item $i$  normalized to the total interactions in training. The corresponding novelty-weighted IDCG would be
\begin{equation}
{\rm Nov\!-\!IDCG}@k = 
\sum_{s=1}^{\mathcal{N}_u(k)} 
\frac{\max_{i\in\mathcal{I}_u}\left(-\ln \nu(i)\right)}
{\log_2(s+1)}.
\end{equation}
In other words, the highest DCG is obtained by ranking the most novel items (among those relevant to the user) in descending order. 
\subsection{Implementation details}\label{subsec:implementation}
The implementation of our model is performed in TensorFlow~\cite{tensorflow2015-whitepaper}.
The model can be trained in both CPU or GPU. 
When GPU is enabled, the use of queues to feed the tensors greatly speeds up the training.
We set the batch size to $100$, and train every DAE model for $120$k iterations, so as to ensure proper convergence. For MF models we use $180$k iterations.
The number of neurons is $200$ in all DAE experiments; for MF models, since the large number of users makes them prone to overfit, we train the models with $100$ and $200$ neurons and take the best performing model.
Weight matrices are initialized with random uniform values whose amplitude is computed as described by Glorot~\emph{et al.}~\cite{Xavier_initialization}. 
For the biases we use a truncated random normal initialization with a standard deviation of $10^{-3}$. 
Models are trained with Adam optimizer~\cite{Kingma2014AdamAM} and a learning rate of $10^{-3}$.

Concerning negative sampling in point and pair--wise schemes, 
we fix the size of the target sets for every user (sets $\mathcal{T}_u$ and $\mathcal{P}_u$ for point and pair--wise learning, respectively, see subsection~\ref{subsec:losses}).
In particular, we make such sets proportional to the median number of items adopted by users, except for the multinomial loss, where all items are utilized~\cite{Liang:2016:CoFactor}.
The proportionality factors are hyper-parameters fine--tuned with the validation set, swapping the values $\{1,\,5,\,10,\,50,\,100,\,150\}$. We find a factor of $50$ or $100$ to provide the best results.


We add noise to the input vector of the AE~\cite{Vincent:2008:ECRF-AE, Wu:2016:CDAE-topN} using drop-out~\cite{liang:2018:VAE}. We fix the level of noise at $0.5$. Competitive performance is achieved after normalizing the AE input vector.
For DAE models, we swap the $L_2$ regularization strength $\lambda\in[10^{-7}-10^{-4}]$, while for MF models we take the form in equation~(\ref{eq:l2_reg_scaled}) with $\lambda\in[10^{0}-10^{3}]$, which provides a more stable training for MF models\footnote{Recall the different scales of the $\lambda$ factor in equations (\ref{eq:L2_reg}) and (\ref{eq:l2_reg_scaled}).}. In general we find that MIL models require smaller $\lambda$ factors than cross-entropy or multinomial--based models. This is expected, as the level of weight--decay regularization in equations~(\ref{eq:L2_reg}) and (\ref{eq:l2_reg_scaled}) depends on the value of the loss, which is smaller for MIL models.

\subsection{Baseline models}\label{subsec:baseline_models}
We implement the objective functions described in subsection \ref{subsec:losses} on a user-based DAE~\cite{Sedhain:2015:Autorec, Wu:2016:CDAE-topN} and compare the results with the MIL function. We also compare them with traditional Matrix Factorization with Weight Regularization~\cite{HuKoren:2008:CF_implicit}. In the following, we provide details on the training of the different models.

\textbf{Weight-Regularized Matrix Factorization} WRMF~\cite{HuKoren:2008:CF_implicit} is a linear factorization model  trained with square loss and weight decay. We use negative sampling with a sampling ratio of $100$ and $\lambda\sim 5-10$ (as obtained in the validation set). We call this model \MFsquare. In addition, we train WRMF models with MIL and point--wise cross--entropy losses, applying a sigmoid function at the output, so as to ensure $\hat{p}_{ui}\in(0,1)$. In these cases, we find that a sampling ratio of $100$ and $\lambda=50-500$ provide best results. We name these models \MFmil\, and \MFce, respectively.

\textbf{\emph{Denoising Autoencoder models}}

\textbf{Cross-entropy loss} For the cross-entropy loss defined in equations~(\ref{eq:cross-entropy}), (\ref{eq:point-wise}) and (\ref{eq:pair-wise}), we use linear--sigmoid and sigmoid--sigmoid activations at the encoder and decoder, respectively. We name the DAEs models with cross-entropy loss and point--wise estimation \CEpointlinsig\,and \CEpointsigsig; and those with pair--wise,  \CEpairlinsig \,and \CEpairsigsig. In order to prevent numerical instabilities, we ensure that the output preferences are in $[\varepsilon, 1-\varepsilon]$, with $\varepsilon=10^{-5}$. Regarding negative sampling, we find that the best sampling ratio is $50\times {\rm median}(\mathcal{I}_u)$ and $100\times {\rm median}(\mathcal{I}_u)$  for point and pair--wise estimation, respectively. Best weight-decay regularization is found to be $\lambda=2\cdot 10^{-5}$.

The closest model to these baselines is the Collaborative Denoising AE (CDAE)~\cite{Wu:2016:CDAE-topN}, although for the sake of simplicity, in the present paper we do not include the user embedding of CDAE. 
Similar to CDAE, we find that  pair--wise learning does not achieve competitive results at the top of the ranked list~\cite{Wu:2016:CDAE-topN, liang:2018:VAE}. 

\textbf{Multinomial loss} AEs trained with a multinomial log-likelihood have recently been  introduced by Lian et al~\cite{liang:2018:VAE}, either applied to DAEs or Variational AEs (VAE) with partial regularization. Here, we focus on the multi-DAE modeling with $\tanh$-linear activations\footnote{
We use the actual implementation provided at \url{https://github.com/dawenl/vae_cf} to verify that the activation used at the decoder of multi-DAE is linear, although the original writing~\cite{liang:2018:VAE} suggests a $\tanh$ non-linearity for the decoder.
}, and name this baseline \MULTItanhlin. Our implementation exactly reproduces that of~\cite{liang:2018:VAE} when using $\lambda=2\cdot 10^{-5}$, input noise of $0.5$ and without applying negative sampling. 

\textbf{Missing Information loss} We apply the \textsc{MIL} function defined in equation~(\ref{eq:mil_def}) to linear-sigmoid and sigmoid-sigmoid DAEs. We name these models \MILlinsig\, and \MILsigsig, respectively. Best hyper-parameters of the loss turn out to be $A_{\rm MI}=10^6,\, \gamma_{\rm MI}=10$ and $\gamma_{+}=1$, after grid search the pairs $\left(A_{\rm MI}, \gamma_{\rm MI}\right)\in\{
(5\cdot10^1, 2)$, $(10^3, 4)$, $(2\cdot10^4, 6)$, $(5\cdot10^5, 10)$, $(1\cdot10^6, 10)$, $(5\cdot10^6, 10)$, $(5\cdot10^9, 15)\}$, and $\gamma_{+}=1$ or $2$. In addition, we use a sampling ratio of $50$ and $\lambda\in(10^{-6}, 10^{-5})$.

\setlength{\belowcaptionskip}{5pt}
\setlength\tabcolsep{3pt}
\begin{table}[hbt]
\begin{tabular}{c c c c c c}
\multicolumn{6}{c}{\textsc{ML20M}} \\
\hline
\hline
  & r@1 & r@20 & n@20 & n@100 & nov@100\\
\hline
\MFsquare & 0.191 & 0.329 & 0.231 & 0.318 & 0.305 \\
\MFce & 0.264 & 0.354 & 0.265 & 0.348 & 0.330 \\
\MFmil & 0.261 & 0.335 & 0.252 & 0.336 & 0.318 \\
\hdashline
\CEpointlinsig & 0.284 & 0.372 & 0.281 & 0.366 & 0.348 \\
\CEpointsigsig & 0.281 & 0.379 & 0.285 & 0.371 & 0.353 \\
\CEpairlinsig & 0.244 & 0.358 & 0.262  & 0.349 & 0.333 \\
\CEpairsigsig & 0.171 & 0.360 & 0.249  & 0.337 & 0.320 \\
\MULTItanhlin & 0.249 & 0.364 & 0.269  & 0.357 & 0.339 \\
\MILlinsig & 0.299 & 0.375 & 0.286  & 0.369 & 0.349 \\
\MILsigsig & 0.301 & 0.373 & 0.285  & 0.367 & 0.348 \\
\hline
\end{tabular}

\caption{Relevance and novelty metrics for the \textsc{ML20M} datasets. Here, $\boldsymbol{{\rm r}@k}$ stands for Recall$\boldsymbol{@k}$, $\boldsymbol{{\rm n}@k}$ is the NDCG$\boldsymbol{@k}$, and $\boldsymbol{{\rm nov}@k}$ the novelty-weighted NDCG$\boldsymbol{@k}$.
The horizontal dash line separate MF models from DAE models. Errors in metrics due to random initialization are $\boldsymbol{\pm 3\cdot10^{-3}}$.
}
\label{table:metrics_results_ml20m}
\end{table}

\section{Experimental Results}\label{sec:results}
\subsection{Performance of Metrics}
Tables~\ref{table:metrics_results_ml20m}, \ref{table:metrics_results_netflix} and \ref{table:metrics_results_lastfm} show the performance of models on the \textsc{ML20M}, \textsc{Netflix} and \textsc{Lastfm} datasets, respectively. The results of the MF and DAE models are shown above and below  the dash lines, respectively. 
Cross-entropy and MIL objective functions applied to MF clearly outperform the traditional \MFsquare\, model in all datasets, being \MFce\,  the one that provides best results among MF models.
Nevertheless, all MF models provide significantly poorer performance than their DAEs counterparts. 

Given the superior performance of DAE over MF models, we focus the rest of the analysis on the former architecture.  
As observed, cross-entropy models in point--wise estimation outperform those in pair--wise estimation to a great extent, in agreement with recent literature~\cite{Steck:2015:GaussianMF, Wu:2016:CDAE-topN, liang:2018:VAE}. Concerning the choice of encoding activations (linear or sigmoid), differences in metric values are in most cases within the error due to random initialization ($\pm 3\cdot 10^{-3}$). However, since \CEpointlinsig\,  outperforms \CEpointsigsig\, in all cases (expect for large top-$k$ ranking in the \textsc{ML20m} dataset), we take \CEpointlinsig\, as the best performing model among those trained with cross-entropy loss (point and pair wise). 

\begin{table}[htb]
\begin{tabular}{c c c c c c}
\multicolumn{6}{c}{\textsc{Netflix}} \\
\hline
\hline
  & r@1 & r@20 & n@20 & n@100 & nov@100\\
\hline
\MFsquare & 0.205 & 0.255 & 0.203 & 0.285 & 0.271 \\
\MFce & 0.262 & 0.278 & 0.230 & 0.308 & 0.292 \\
\MFmil & 0.269 & 0.272 & 0.227 & 0.302 & 0.284 \\
\hdashline
\CEpointlinsig & 0.300 & 0.305 & 0.256 & 0.334 & 0.319 \\
\CEpointsigsig & 0.283 & 0.304 & 0.251 & 0.332 & 0.317 \\
\CEpairlinsig & 0.247 & 0.283 & 0.230 & 0.311 & 0.297 \\
\CEpairsigsig & 0.255 & 0.279 & 0.226 & 0.307 & 0.292 \\
\MULTItanhlin & 0.241 & 0.290 & 0.234 & 0.320 & 0.305 \\
\MILlinsig & 0.304 & 0.304 & 0.256 & 0.332 & 0.316 \\
\MILsigsig & 0.304 & 0.299 & 0.252 & 0.328 & 0.311 \\
\hline
\end{tabular}
\caption{Relevance and novelty metrics for the \textsc{Netflix} dataset. See  Table~\ref{table:metrics_results_ml20m} caption for details.
}
\label{table:metrics_results_netflix}
\end{table}

Regarding MIL models, they are in pair (within statistical errors) with cross-entropy models in point wise estimation. There is a tendency for MIL to perform better at low top-$k$, while cross-entropy seems to outperform MIL at large top-$k$ rankings. However, such differences are quite small, and could be easily attributed to other source of systematic errors. With respect to the choice of encoding activations within MIL--based models, as in the case of cross-entropy, linear encoders perform better (without statistical significance, though).
On the other hand, the multinomial log-likelihood does not achieve good performance at low top-$k$. Indeed, the reported metric values at $k\leq20$ are closer to those of pair--wise models than to the best performing models (\CEpointlinsig\, and \MILlinsig). Nonetheless, \MULTItanhlin\, model metric values  recover at large $k$. 

\begin{table}[htb]
\begin{tabular}{c c c c c c}
\multicolumn{6}{c}{\textsc{lastfm}} \\
\hline
\hline
  & r@1 & r@20 & n@20 & n@100 & nov@100\\
\hline
\MFsquare & 0.155 & 0.236 & 0.172 & 0.243 & 0.228 \\
\MFce & 0.175 & 0.258 & 0.191 & 0.265 & 0.251 \\
\MFmil  & 0.163 & 0.242 & 0.178 & 0.250 & 0.236 \\
\hdashline
\CEpointlinsig & 0.228 & 0.300 & 0.229 & 0.305 & 0.289 \\
\CEpointsigsig & 0.218 & 0.293 & 0.222 & 0.299 & 0.283 \\
\CEpairlinsig & 0.153 & 0.277 & 0.193 & 0.275 & 0.261 \\
\CEpairsigsig & 0.184 & 0.279 & 0.204 & 0.284 & 0.270 \\
\MULTItanhlin & 0.187 & 0.287 & 0.210 & 0.289 & 0.275 \\
\MILlinsig & 0.222 & 0.296 & 0.225 & 0.299 & 0.283 \\
\MILsigsig & 0.222 & 0.294 & 0.224 & 0.299 & 0.283 \\
\hline
\end{tabular}
\caption{Relevance and novelty metrics for the \textsc{LAstfm} datasets. See  Table~\ref{table:metrics_results_ml20m} caption for details.
}
\label{table:metrics_results_lastfm}
\end{table}
\setlength\tabcolsep{6pt}

In order to further establish the relative performance of the multinomial loss, we present in Table~\ref{table:metrics_results_vae} the results for the data processing presented in~\cite{liang:2018:VAE} for the \textsc{ML20m} dataset\footnote{
We use the \textsc{ml20m} dataset, since is the only one provided in their implementation, see \url{https://github.com/dawenl/vae_cf}.
}, in which the test set consist of held out users. For this, we use their own implementation, slightly changed so that DAE can be trained with MIL and cross-entropy objective functions as well. The results for the models above the dash line in Table~\ref{table:metrics_results_vae} are taken directly from~\cite{liang:2018:VAE}. As observed, \textsc{Multi-dae} (i.e. \MULTItanhlin\, in our nomenclature) metric values are in pair (for recalls at large top-$k$) or below (for small top-$k$, or when accounting for the ranking order, as in NDCG) the performance of MIL and cross-entropy based DAEs. This is in agreement with the conclusions drawn from the experimental results in Tables~\ref{table:metrics_results_ml20m}, \ref{table:metrics_results_netflix} and \ref{table:metrics_results_lastfm}. 
Please note that the NAs in Table~\ref{table:metrics_results_vae} stand for \emph{not available} results, because they were not reported in ~\cite{liang:2018:VAE} and the code used is not publicly available.
For completeness, we include in Table~\ref{table:metrics_results_vae} the results of the Variational AE (VAE)~\cite{liang:2018:VAE}, \textsc{Multi-VAE}$^{\rm PR}$, which are indeed close to those obtained with MIL and cross-entropy. Nevertheless, a proper comparison with \textsc{Multi-VAE}$^{\rm PR}$ (which requires the implementation of other losses for VAE) is out of the scope of this work. 

\begin{table}[htb]
\begin{tabular}{c c c c c c}
\multicolumn{6}{c}{\textsc{ML20m held-out test}} \\
\hline
\hline
  & r@1 & r@20 & r@50 & n@20 & n@100 \\
\hline
\vspace{-0.3cm}
\\
\textsc{Multi-VAE}$^{\rm PR}$ & 0.378 & 0.395 & 0.537 & 0.336 & 0.426 \\
\textsc{Multi-DAE} & 0.383 & 0.387 & 0.524 & 0.331 & 0.419 \\
\textsc{WMF} & NA & 0.360 & 0.498 & NA & 0.386 \\
\textsc{SLIM} & NA & 0.370 & 0.495 & NA & 0.401  \\
\textsc{CDAE} & NA &0.391 & 0.523 & NA & 0.418 \\
\hdashline
\CEpointlinsig & 0.404 & 0.389 & 0.518 & 0.338 & 0.419 \\
\CEpointsigsig & 0.401 & 0.395 & 0.532 & 0.343 & 0.428 \\
\MILlinsig & 0.409 & 0.392 & 0.520 & 0.341 & 0.421 \\
\MILsigsig & 0.411 & 0.394 & 0.526 & 0.343 & 0.425 \\
\hline
\end{tabular}
\caption{Relevance and novelty metrics for the \textsc{ML20m} dataset as process in \cite{liang:2018:VAE}. 
The horizontal dash line separate models reported in \cite{liang:2018:VAE} from those calculated in this work following their data process and implementation. NA stand for \emph{not available} results, because they were not reported in ~\cite{liang:2018:VAE} and the code used is not publicly available. Here, $\boldsymbol{{\rm r}@k}$ stands for Recall$\boldsymbol{@k}$ and  $\boldsymbol{{\rm n}@k}$ is the NDCG$\boldsymbol{@k}$.
}
\label{table:metrics_results_vae}
\end{table}

Given the similarity of \textsc{MIL} and \textsc{CE$_{\rm Point}$} DAE models in terms of relevance-aware metrics, we proceed next to study the differences in the distribution of recommendations, and how are these allocated in terms of the popularity distribution of the items. 

\subsection{Distribution of preferences}
In this sub-section, we study the distribution of predicted user preferences by DAE models averaged across all users, $\langle \hat p\rangle$. We focus on objective functions modeling preferences; the \textsc{MULTI} model deals with probabilities across a multi-class problem, and thus cannot be easily compared. Table~\ref{tab:comparison_distributions} presents the user--averaged distribution of predicted preferences by the \CEpointlinsig, \CEpairlinsig\, and \MILlinsig\, models, for the \textsc{Netflix} dataset. As observed, traditional losses tend to set most items with a small preference, close to zero, as expected from the cross--entropy loss. Yet, there is a clear distinction between point-- and pair--wise learning. On average, the point-wise cross-entropy model tends to set few items with  high  preference for each user (fewer than $100$ items with $\langle p\rangle\ge0.9$), while recommending more than $85~\%$ of the available catalogue with an almost zero preference. On the other hand, pair-wise models set a considerably higher proportion of items with a measurable preference (around 5 times larger). Hence, 
\textsc{CE point} optimizes the head of the recommendation by setting very few items with high preference for each user; on the other hand, \textsc{CE pair} allows more items to have a high preference in the recommendation, which may cause a less effective optimization of the ranked list (in agreement with the results presented in Tables~\ref{table:metrics_results_ml20m}, \ref{table:metrics_results_netflix} and \ref{table:metrics_results_lastfm}). 

\begin{table}[h]
\begin{tabular}{c c c | c c c}
\multicolumn{6}{c}{\textsc{Distribution of predicted preferences}}
\\
\hline 
\hline
\multicolumn{3}{c}{preference thresholds} & \textsc{CE point} & \textsc{CE pair} & \textsc{MIL} \\
\hline 
1.0 & $\ge \langle\hat{p}\rangle \ge $ & 0.9   & 0.35 & 1.62 & 2.32 \\
0.9 &$ > \langle\hat{p}\rangle \ge $ & 0.7     & 0.33 & 1.58 & 60.2 \\
0.7 &$ > \langle\hat{p}\rangle \ge $ & 0.5     & 0.35 & 2.21 & 31.6 \\
0.5 &$ > \langle\hat{p}\rangle \ge $ & 0.25    & 0.79 & 6.49 & 5.49 \\
0.25 &$ > \langle\hat{p}\rangle \ge $ & 0.01   & 13.1 & 49.8 & 0.39 \\
0.01 &$ > \langle\hat{p}\rangle \ge $ & 0.0    & 85.1 & 38,3 & 0.0 \\
\hline
\end{tabular}
\caption{Comparison of the distribution of predicted preferences
averaged across all users in the \textsc{Netflix} dataset, $\boldsymbol{\langle\hat{p}\rangle}$, 
for \textsc{CE point}, \textsc{CE pair} and \textsc{MIL} models with  linear--sigmoid activations.}
\label{tab:comparison_distributions}
\end{table}

Conversely, the MIL function pushes all items towards high preferences. This is a consequence of the functional form in equation~(\ref{eq:mil_def}), where for large $\gamma_{\rm MI}$ the missing information term does not contribute to the loss unless the predicted preference is close to $1$ or $0$. Thus, the ranking of unseen items is left to the low--rank process, rather than forcing unobserved items to be at the tail of the recommendation ($\hat{p}=0$). Such a ranking optimization has an important consequence: it allows all items to have a chance to be recommended, since none of them have a zero preference prediction. This effect might be of interest for RS services that cannot recommend all the items in their catalogue--due to legal constraints, for instance, or because of some particular business requirements.

\subsection{Popularity distribution of the recommendations}
\begin{table}[htb]
\begin{tabular}{c c c}

\hline
\hline
  & Short $\rightarrow$ Medium & Medium $\rightarrow$ Long \\
\hline
\textsc{ml20m} & 177 & 784 \\
\textsc{Netflix} & 251 & 965 \\
\textsc{lastfm} & 351 & 2240 \\
\end{tabular}
\caption{
Interval cuts of the popularity distribution of items. 
}
\label{table:intervals_long_tail}
\end{table}

The question of what kind of items (\emph{i.e.}, popular, frequent or infrequent) are recommended by each model is yet to be answered. To this end, we examine how the top-200 recommendations are distributed on the short, medium and long--tail intervals of the popularity distribution. 
Inspired by Celma \emph{et al.}~\cite{Celma:2008:approach_novelty}, we calculate the cumulative distribution of item adoptions, $F(x)$, and take the short--tail interval as composed by the first $N_{33}$ items, where $N_{33}$ is the number of items that covers one third of the total visualizations, \emph{i.e.} $F(N_{33})=33 \%$. Similarly, the medium--tail items account for the second third of the total visualizations, \emph{i.e.} items in $(N_{33}, N_{66}]$ with $F(N_{66})=66 \%$. The rest of the item catalogue is taken within the long--tail interval. 
Table~\ref{table:intervals_long_tail} depicts the resulting interval cuts for each dataset. As observed, the \textsc{lastfm} dataset has the most heavy--tailed distribution (its short and medium tail contain the larger amount of items among the datasets used in this work), while \textsc{ml20m} presents the less. 

\begin{table}[htb]
\begin{tabular}{c c c c}

\multicolumn{4}{c}{\textsc{ml20m}} \\
\hline
\hline
  & Short & Medium &  Long \\
\hline
\CEpointlinsig & 41.1 & 40.1 & 18.8 \\
\MULTItanhlin & 36.3 & 40.8 & 22.9 \\
\MILlinsig & 32.2 & 39.9 & 27.9 \\
\\

\multicolumn{4}{c}{\textsc{netflix}} \\
\hline
\hline
  & Short & Medium &  Long \\
\hline
\CEpointlinsig & 52.6 & 34.2 & 13.2 \\
\MULTItanhlin & 47.7 & 36.0 & 16.3 \\
\MILlinsig & 41.7 & 37.8 & 20.5 \\
\\

\multicolumn{4}{c}{\textsc{lastfm}} \\
\hline
\hline
  & Short & Medium &  Long \\
\hline
\CEpointlinsig & 39.6 & 34.8 & 25.6 \\
\MULTItanhlin & 36.0 & 35.0 & 29.0 \\
\MILlinsig & 33.4 & 37.2 & 29.4 \\
\\

\end{tabular}
\caption{
Distribution of top-200 recommendations (in percentage) within popularity intervals. 
}
\label{table:percentages_popularity}
\end{table}

Table~\ref{table:percentages_popularity} presents the distribution of top-200 recommendations within the  popularity intervals just defined, as obtained with \CEpointlinsig, \MULTItanhlin\, and \MILlinsig. 
Not surprisingly, \CEpointlinsig\, achieves large metric values by being the model that recommends popular items most frequently in all datasets. 
This popularity bias in the recommendation is alleviated by the multinomial loss, concomitant to an incremental decrease of metric values (mainly at small top-$k$ rankings), as demonstrated in Tables~\ref{table:metrics_results_ml20m}, \ref{table:metrics_results_netflix} and \ref{table:metrics_results_lastfm}. 
Remarkably, \MILlinsig\, yields metric performances similar to those of \CEpointlinsig\,  while recommending popular items less frequently. For instance, in the \textsc{ml20m} and \textsc{Netflix} datasets, \MILlinsig\, recommend popular items $10$ percentage points less than the \CEpointlinsig\, model (a $\sim20 \%$ decrease in both datasets). For the \textsc{lastfm} dataset, the decrease is of $6$ percentage points ($15 \%$ decrease). Moreover, such a  decrease in short--tail recommendations  favours the appearance of both medium and long--tail items at the top-200 list. For instance, the \MILlinsig\, model recommends long--tail items $\sim50~\%$ more frequently than the \CEpointlinsig\, model for the \textsc{ml20m} and \textsc{Netflix} datasets. 
On the other hand, the heavier tail of the \textsc{Lastfm} dataset makes the recommendations of all models to be more evenly distributed among the intervals of popularity. Still, \MILlinsig\, continues to be the model that recommends popular items less frequently. 

\begin{figure}
\begin{center}
    \includegraphics[width=.9\linewidth]{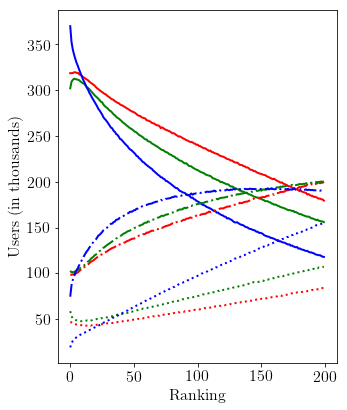}
    \caption{Ranking evolution of short (solid lines), medium (dash lines) and long--tail (dots) items within the top-200 recommendations for the \textsc{Netflix} dataset.
    \MILlinsig\, model is presented in blue lines, \CEpointlinsig\, in red, and \MULTItanhlin\, in green.}
    \label{fig:rankings_tails_netflix}
\end{center}
\end{figure}

We finally analyze in Figure~\ref{fig:rankings_tails_netflix} how items belonging to the short (solid lines), medium (dash lines) or long--tail (dots) popularity intervals are ranked within the top-200 recommendations, for the \textsc{Netflix} dataset. 
As observed, the \CEpointlinsig\, model attacks mainly short--tail items for ranking positions smaller than $100$ (red solid line). This explains the competitive metric scores reported in Table~\ref{table:metrics_results_netflix}, as well as the relatively few items set with a high preference, see Table~\ref{tab:comparison_distributions}. 
On the other hand, the \textsc{Multi} (green) model reduces the number of popular items ranked in the top positions, while recommending more items from the medium and long--tail. This change from short-- towards medium and long--tail recommendations may explain the relatively poorer performance of the multinomial loss at small top-$k$ values.

The \CEpointlinsig\, model tends to under-represent infrequent items (red dots).
This misrepresentation of long--tail items is alleviated by \textsc{MIL} models.
They heavily set short-tail items at the very top of the ranking (blue solid lines), but drastically reduce their appearance shortly thereafter. Instead, \textsc{MIL} models sharply increase the recommendation of medium--tail items (blue dash lines)  until the top-100 ranking, where the growth rate stagnates. Meanwhile, for items belonging to the long--tail (blue dots), \textsc{MIL} expands its appearance almost linearly, to the point that they exceed the number of short--tail recommendations. 
We finally highlight  that for ranking position $100$, \MILlinsig\, recommends long--tail items to $\sim40k$ users more than \CEpointlinsig\, and $\sim25k$ users more than \MULTItanhlin. 


\section{Conclusions and next steps}\label{sec:conclusions}

In this paper we present a novel objective function, the \emph{Missing Information Loss} (MIL), specifically designed for handling unobserved user-item interactions in implicit feedback datasets. In particular, MIL explicitly forbids treating missing user-item interactions as positive or negative feedback.
We demonstrate that, thanks to the functional form of the MIL function, the ranking of unseen items is almost entirely left to the low--rank process, rather than forcing unobserved items to be at the tail of the recommendation (\emph{i.e.}, MIL does not force a zero predicted preference for unobserved user-item interactions). 

Extensive experiments with Matrix Factorization and Denoising Autoencoders conducted on three datasets, show that \textsc{MIL} models demonstrate competitive performance when compared with other traditional losses such as cross-entropy or the multinomial log-likelihood. 
In addition, we study the distribution of the recommendations and observe that the reported metric performance takes place while recommending popular items less frequently (up to a $20 \%$ decrease with respect to the best competing method). Indeed, \textsc{MIL} models sharply increase the recommendation of medium--tail items, while almost linearly expanding the appearance of long--tail items with the ranking position in the list of recommendations. Such expansion results in up to a $50 \%$ increase of long--tail recommendations, a feature of utmost importance for industries with a large catalogue of items. 

Future lines of research may involve the incorporation of negative feedback, or the usage of \textsc{MIL} in temporal--aware Recommender Systems (such as those using Recurrent Neural Networks).  
In addition, we hope that the results here reported  will bring forward first-principle mathematical derivations of the \textsc{MIL} function, so that the vast family of possible polynomials modelling the missing information term can be reduced, or even extended with more suitable functions. 

\begin{acks}
We would like to thank the continuous support and careful reading of the manuscript by the \emph{Edge} guild within BBVA Data \& Analytics, specially J. Garc\'ia Santamar\'ia and J. A. Rodr\'iguez Serrano. 
\end{acks}

\bibliographystyle{unsrt}

\begin{thebibliography}{10}

\bibitem{Anderson:2006:long_tail}
Chris Anderson.
\newblock {\em The Long Tail: Why the Future of Business Is Selling Less of
  More}.
\newblock Hyperion, 2006.

\bibitem{Steck:2011:IPR}
Harald Steck.
\newblock Item popularity and recommendation accuracy.
\newblock In {\em Proceedings of the Fifth ACM Conference on Recommender
  Systems}, RecSys '11, pages 125--132, New York, NY, USA, 2011. ACM.

\bibitem{HuKoren:2008:CF_implicit}
Yifan Hu, Yehuda Koren, and Chris Volinsky.
\newblock Collaborative filtering for implicit feedback datasets.
\newblock In {\em 2008 Eighth IEEE International Conference on Data Mining},
  pages 263--272, Dec 2008.

\bibitem{Pan:2008:OCCF}
Rong Pan, Yunhong Zhou, Bin Cao, Nathan~N. Liu, Rajan Lukose, Martin Scholz,
  and Qiang Yang.
\newblock One-class collaborative filtering.
\newblock In {\em Proceedings of the 2008 Eighth IEEE International Conference
  on Data Mining}, ICDM '08, pages 502--511, Washington, DC, USA, 2008. IEEE
  Computer Society.

\bibitem{Steck:2010:MissingNotAtRandom}
Harald Steck.
\newblock Training and testing of recommender systems on data missing not at
  random.
\newblock In {\em Proceedings of the 16th ACM SIGKDD International Conference
  on Knowledge Discovery and Data Mining}, KDD '10, pages 713--722, New York,
  NY, USA, 2010. ACM.

\bibitem{liang:2018:VAE}
Dawen Liang, Rahul~G. Krishnan, Matthew~D. Hoffman, and Tony Jebara.
\newblock Variational autoencoders for collaborative filtering, 2018.
\newblock cite arxiv:1802.05814Comment: 10 pages, 3 figures. WWW 2018.

\bibitem{Vincent:2008:ECRF-AE}
Pascal Vincent, Hugo Larochelle, Yoshua Bengio, and Pierre-Antoine Manzagol.
\newblock Extracting and composing robust features with denoising autoencoders.
\newblock In {\em Proceedings of the 25th International Conference on Machine
  Learning}, ICML '08, pages 1096--1103, New York, NY, USA, 2008. ACM.

\bibitem{Wu:2016:CDAE-topN}
Yao Wu, Christopher DuBois, Alice~X. Zheng, and Martin Ester.
\newblock Collaborative denoising auto-encoders for top-n recommender systems.
\newblock In {\em Proceedings of the Ninth ACM International Conference on Web
  Search and Data Mining}, WSDM '16, pages 153--162, New York, NY, USA, 2016.
  ACM.

\bibitem{Kramer:1991:NLPCA}
Mark~A. Kramer.
\newblock Nonlinear principal component analysis using autoassociative neural
  networks.
\newblock {\em AIChE Journal}, 37(2):233--243, 1991.

\bibitem{Sedhain:2015:Autorec}
Suvash Sedhain, Aditya~Krishna Menon, Scott Sanner, and Lexing Xie.
\newblock Autorec: Autoencoders meet collaborative filtering.
\newblock In {\em Proceedings of the 24th International Conference on World
  Wide Web}, WWW '15 Companion, pages 111--112, New York, NY, USA, 2015. ACM.

\bibitem{Rendle:2009:BPR}
Steffen Rendle, Christoph Freudenthaler, Zeno Gantner, and Lars Schmidt-Thieme.
\newblock Bpr: Bayesian personalized ranking from implicit feedback.
\newblock In {\em Proceedings of the Twenty-Fifth Conference on Uncertainty in
  Artificial Intelligence}, UAI '09, pages 452--461, Arlington, Virginia,
  United States, 2009. AUAI Press.

\bibitem{icml2010_NairH10_Relu}
Vinod Nair and Geoffrey~E. Hinton.
\newblock Rectified linear units improve restricted boltzmann machines.
\newblock In Johannes Fürnkranz and Thorsten Joachims, editors, {\em
  Proceedings of the 27th International Conference on Machine Learning
  (ICML-10)}, pages 807--814. Omnipress, 2010.

\bibitem{Celma:Springer2010}
O.~Celma.
\newblock {\em {Music Recommendation and Discovery in the Long Tail}}.
\newblock Springer, 2010.

\bibitem{Vargas:2011:Novelty_diversity}
Sa\'{u}l Vargas and Pablo Castells.
\newblock Rank and relevance in novelty and diversity metrics for recommender
  systems.
\newblock In {\em Proceedings of the Fifth ACM Conference on Recommender
  Systems}, RecSys '11, pages 109--116, New York, NY, USA, 2011. ACM.

\bibitem{tensorflow2015-whitepaper}
Mart\'{\i}n Abadi, Ashish Agarwal, Paul Barham, Eugene Brevdo, Zhifeng Chen,
  Craig Citro, Greg~S. Corrado, Andy Davis, Jeffrey Dean, Matthieu Devin,
  Sanjay Ghemawat, Ian Goodfellow, Andrew Harp, Geoffrey Irving, Michael Isard,
  Yangqing Jia, Rafal Jozefowicz, Lukasz Kaiser, Manjunath Kudlur, Josh
  Levenberg, Dan Man\'{e}, Rajat Monga, Sherry Moore, Derek Murray, Chris Olah,
  Mike Schuster, Jonathon Shlens, Benoit Steiner, Ilya Sutskever, Kunal Talwar,
  Paul Tucker, Vincent Vanhoucke, Vijay Vasudevan, Fernanda Vi\'{e}gas, Oriol
  Vinyals, Pete Warden, Martin Wattenberg, Martin Wicke, Yuan Yu, and Xiaoqiang
  Zheng.
\newblock {TensorFlow}: Large-scale machine learning on heterogeneous systems,
  2015.
\newblock Software available from tensorflow.org.

\bibitem{Xavier_initialization}
Xavier Glorot and Yoshua Bengio.
\newblock Understanding the difficulty of training deep feedforward neural
  networks.
\newblock In {\em International conference on artificial intelligence and
  statistics}, AISTATS 2010, pages 249--256, 2010.

\bibitem{Kingma2014AdamAM}
Diederik~P. Kingma and Jimmy Ba.
\newblock Adam: A method for stochastic optimization.
\newblock {\em CoRR}, abs/1412.6980, 2014.

\bibitem{Liang:2016:CoFactor}
Dawen Liang, Jaan Altosaar, Laurent Charlin, and David~M. Blei.
\newblock Factorization meets the item embedding: Regularizing matrix
  factorization with item co-occurrence.
\newblock In {\em Proceedings of the 10th ACM Conference on Recommender
  Systems}, RecSys '16, pages 59--66, New York, NY, USA, 2016. ACM.

\bibitem{Steck:2015:GaussianMF}
Harald Steck.
\newblock Gaussian ranking by matrix factorization.
\newblock In {\em Proceedings of the 9th ACM Conference on Recommender
  Systems}, RecSys '15, pages 115--122, New York, NY, USA, 2015. ACM.

\bibitem{Celma:2008:approach_novelty}
\`{O}scar Celma and Perfecto Herrera.
\newblock A new approach to evaluating novel recommendations.
\newblock In {\em Proceedings of the 2008 ACM Conference on Recommender
  Systems}, RecSys '08, pages 179--186, New York, NY, USA, 2008. ACM.

\end{thebibliography}

\end{document}